# Ontological Foundations of State Sovereignty

John Beverley
*Department of Philosophy*
*University at Buffalo*
Buffalo, USA
0000-0002-1118-1738

Danielle Limbaugh
*Department of Philosophy*
*Cornell University*
Ithaca, USA

*Abstract—* **This short paper is a primer on thenature of state sovereignty and the importance of claims about it. It also aims to reveal (*merely* reveal) a strategy for working with vague or contradictory data about which states, in fact, are sovereign. These goals together areintended to set the stage for applied work in ontology about international affairs.**

## INTRODUCTION

This short paper is a primer on the nature of state sovereignty and the importance of claims about it. It also aims to reveal (*merely* reveal) a strategy for working with vague or contradictory data about which states, in fact, are sovereign. These goals together are intended to set the stage for applied work in ontology about international affairs.

State sovereignty has long stood at the core of political theory, playing a pivotal role in determining the relationships between geopolitical entities and citizens. State sovereignty is, moreover, deeply rooted in normative and social frameworks that shape how states interact on the global stage. The goal of this paper is to provide a clarifying ontological theory of state sovereignty that is sensitive to the significance of United Nations (UN) declarations, but which also can afford international disagreement and competing claims. This work provides a strategy and an ontological foundation for modeling the theoretical and normative underpinnings exhibited in the domain of state sovereignty, including (perhaps most importantly) geopolitical claims about sovereignty. It provides a way for scholars, policymakers, and legal practitioners to navigate the complex intersections of state sovereignty, self-determination, and state recognition.

The paper proceeds as follows: The next section (§2) highlights the complexity of state sovereignty. Examining data (mostly unstructured) on state sovereignty reveals why an ontological foundation is essential for addressing current challenges. Relevant sources of international law are then explored (§3) to help discover the features of state sovereignty (§4). These features pave the way for the formal modeling of state sovereignty (§5). Finally, §6 concludes.

## THE COMPLEXITY OF STATE SOVREIGNTY

L.F.L. Oppenheim sheds light on the intricate and contentious nature of state sovereignty when he says,

> "There exists perhaps no conception the meaning of which is more controversial than that of state sovereignty. It is an indisputable fact that this conception, from the moment when it was introduced into political science until the present day, has never had a meaning, which was universally agreed upon" [1].

The complexity of state sovereignty arises from at least two reasons. First, international law holds immense importance in shaping the global order, yet it remains inherently under-described. Unlike domestic legal systems within nations, where centralized authorities enforce laws and provide definitive interpretations, the international sphere lacks a universal jurisdiction with comparable authority. This absence of a cohesive enforcement mechanism or interpretative body results in ambiguity and inconsistency, leaving room for varied interpretations and disputes. Second, discussions about state sovereignty often result in miscommunication because the term "state sovereignty" is used in multiple ways. For instance, it can refer to normative beliefs, such as whether a particular state *should* or *should not* be sovereign. Normative beliefs such as this are widespread in the Israeli-Palestinian conflict [2]. Alternatively, it may describe the functional reality of sovereignty, such as whether a state *does* or *does not* operate as a sovereign entity. For example, while Ukraine is considered a sovereign state, its ability to exercise sovereignty is limited due to Russia's external interference [3].

Despite this complexity, it is evident that entities like the United States, Canada, and the United Kingdom differ in some fundamental way from entities such as Antarctica or the Commonwealth of Puerto Rico. We maintain that one difference between these groups is what we refer to as "state sovereignty," which the former undeniably exhibit and the latter do not. In what follows, we provide first steps towards an ontological characterization of this phenomena. This is not merely a theoretical issue but is indeed a practical one, as sovereignty underpins critical aspects of the global order, such as entering treaties, resolving conflicts, engaging in trade, and maintaining diplomatic relations. Sovereignty plays a crucial role in shaping geopolitical affairs, influencing how states interact with one another. It is thus important that we have a robust understanding of the phenomenon.

Given the complexities of state sovereignty, it is no surprise that organizing and structuring data related to state sovereignty raises its own difficulties. Data surrounding state sovereignty is sometimes vague or contradicts itself, making it challenging to draw definitive conclusions. Consider Taiwan who considers itself sovereign and properly referred to as 'The Republic of China' [4]. The Geopolitical Entities, Names, and Codes Standard (GENC) implies Taiwan independence by not recording Taiwan as part of any other geopolitical entity. Yet, GENC also denies Taiwan its constitutionally self-given full name referring to it simply as 'Taiwan' [6]. The effect is to merely imply that Taiwan is independent but without endorsing Taiwan's self-made claims about the nature of this independence; thus, implying independence without necessarily implying Taiwan's sovereignty [6]. Furthermore, consider conflicting claims made by China and Paraguay; China explicitly denies, while Paraguay explicitly endorses, Taiwan's sovereignty. Modeling the implications of these claims results in three incompatible realities about Taiwan. 1) Taiwan is sovereign, 2) Taiwan is not sovereign, and 3) it is unknown whether Taiwan is sovereign. If we intend to capture the above in a logically coherent structured format, then these realities must be reconciled. Unfortunately, unlike the size or population of a country, we cannot easily (if at all) empirically measure sovereignty to add clarity or establish the veracity of claims. Another solution is required.

We propose paying special attention to claims about sovereignty. Two initial observations. First, while one state may endorse some state's sovereignty while another state denies it, these competing claims about a state's sovereignty vary in significance. Not all claims on state sovereignty are equal. Second, while whether some state is sovereign may be contested, the claims themselves are not. For example, it is universally acknowledged that Taiwan *claims* sovereignty and China denies it; what is at issue is which of these claims is veridical. Preserving data about *who* is making specific claims on state sovereignty and *when* they are made is a strategy for tracking debates and motivations within this area.

## SOURCES OF STATE SOVREIGNTY

The importance of state sovereignty is recognized in membership to the United Nations (UN) [7]. Alongside requiring unanimous approval by the Security Council and a two-thirds majority vote in favor, UN membership also requires agreeing to the UN Charter and its principle of *sovereign equality* among all [8]. Achieving international recognition as a sovereign state is not always straightforward, as it is often influenced by power dynamics and the prevailing international political climate. Nevertheless, once a territory *is* considered a sovereign state by the UN, the UN argues that no outside actor has the right to intervene, directly or indirectly, in the internal or external affairs of a sovereign state [9].

State sovereignty serves as the cornerstone of UN membership because the UN accepts the right to self-determination, roughly, the right to determine, to some extent, their own destiny. For example, the UN International Covenant of Civil and Political Rights reads: "All peoples have the right of self-determination. By virtue of that right they freely determine their political status and freely pursue their economic, social and cultural development" [10]. Similarly, the right to self-determination is further enshrined in the UN Charter: "the principle of equal rights and self-determination of peoples" [11]. Self-determination is widely considered by legal scholars as the basis for justifying state sovereignty [12].

The right to self-determination is closely tied to sovereign statehood because state sovereignty is *a means by which self-determination can be exercised*. State sovereignty *protects* the right to self-determination and fosters an environment conducive to its exercise, enabling groups to exercise rights to determine their political, economic, and social future. Furthermore, the public and explicit recognition of sovereignty by some sufficiently powerful international community – like the UN – is essential for the realization of self-determination, as it grants the legal and political framework within which a people can shape their own destiny. Consider, if a state were constantly subject to outside intervention, it would be unclear how self-determination could be achieved. UN recognition of sovereignty is a means by which such intervention is deterred. Self-determination requires the freedom to make decisions regarding a state's political, economic, and social affairs without external coercion or influence. Continuous outside interference undermines this autonomy.

Relatedly, sovereign statehood thus described requires only *sufficient* authority and *sufficient* independence, as opposed to *absolute* authority and *absolute* independence. The absolutist view of state sovereignty holds that states are sovereign when they have absolute authority over domestic affairs and absolute independence from external actors. The unchecked power granted by the absolutist view may facilitate the oppression of its populace with impunity, thereby halting any ongoing self-determination in the territory. For example, under the absolutist view, the unbridled power bestowed upon the sovereign state allows it to disregard international norms of *jus cogens*. A norm of *jus cogens* is a peremptory norm accepted and recognized by the international community as a norm from which no derogation is allowed and which can only be modified by a subsequent norm of international law having the same character [13]. There are norms of *jus cogens* against slavery, human trafficking, genocide, waging wars of aggression, and crimes against humanity. A state exhibiting absolute state sovereignty would possess the unchecked power to traffic or enslave its citizens, directly impeding any self-determination. The absolutist view is *not* a means by which self-determination can be exercised, rather it is a means by which self-determination can be destroyed.

On the other hand, a limited view of state sovereignty is justified by the right to self-determination [14]. Such a view of state sovereignty acknowledges constraints on a state's authority and independence. This could result from factors such as international treaties, supranational organizations, or agreements with other states that restrict certain aspects of the state's autonomy. Limited sovereignty implies that a state's freedom of action is circumscribed by external factors, imposing restrictions on its ability to govern and interact with other states autonomously. It is this more limited notion of state sovereignty that the UN has in mind, and the notion that will occupy us here.

FEATURES OF STATE SOVREIGNTY

Crucial features of state sovereignty that emerge from both the legal text discussed above and philosophical legal theory discussed below include:

1. Should exhibit sufficient internal authority over territories;
2. Should exhibit sufficient independence from international community; and
3. Requires an act of recognition.

Regarding the first two features, state sovereignty has two main aspects: an internal and external component. These components are not distinct types of state sovereignty but are rather complementary, coexisting aspects [15]. Internal components of sovereignty are a state's sufficient *authority* and control over its domestic affairs within its territorial boundaries [16]. This includes, among other things, the ability to enact and enforce laws, maintain order, provide public services, and govern its population without external interference. It encompasses features such as the establishment of government institutions, the administration of justice, and the regulation of economic activities within the state. Importantly, this internal component is normative. Sovereign states *should* be legitimate governing authorities, although they may not be. Legitimate authority is in contrast to coercive power. A state can have coercive power to command and control without it being a legitimate governing power. The statement "I obey your command out of fear for my life, but I do not acknowledge your claim to authority" is coherent. Of course, the distinction between authority and mere power does not imply that they are unrelated. It may be that sovereignty cannot be had without some power. As Philpott says, "If sovereignty is not mere power, neither is it mere legitimacy" [17].

Regarding the second feature, external components of sovereignty are a state's sufficient *independence* and autonomy in its interactions with other states and international actors [18]. It involves the state's recognition as a legal and political entity by other states and its entitlement to establish and maintain foreign relationships, enter into treaties and agreements, and represent its interests on the global stage. External components of sovereignty promote an environment conducive to exercising self-determination. It also entails the protection of territorial integrity and defense against external threats or aggression. Independence is also normative. Outsider actors in the international community *should not* interfere in the affairs of sovereign states. This allows the sovereign state to exercise its authority within its territorial boundaries. However, the normative nature of independence makes sovereign independence distinct from the mere ability to act independently. For example, Ukraine is recognized as a sovereign state, meaning that outside actors should respect its autonomy and refrain from interference in its internal affairs. However, despite its normative sovereignty, Ukraine's practical ability to act independently within its territory is deeply constrained by Russia's interference [19].

The upshot of this is that state sovereignty indicates what *should be* the case not necessarily what *is* the case. This means that while state sovereignty is meant to promote or protect a people's right to self-determination, this does not imply that the peoples of a sovereign state are in fact able to be self-determined. In fact, a state in duress may still be a sovereign state. For example, even though in 2024 Haiti is in duress, it still has the recognition needed by the UN to be considered a sovereign state [20]. In this way state sovereignty is trivially like a right to self-defense; a right to self-defense is meant to promote and protect self-determination at an individual level, and yet one can have a right to self-defense without the capability to exercise that right.

While even states in duress can be sovereign, when a state does not have enough authority and/or independence, questions about the state's sovereignty may emerge. This implies that there may be circumstances in which the justification for state sovereignty can be overridden. To help motivate when justifications for state sovereignty may be overridden, we can compare state sovereignty with parental authority. Just as there are good reasons for parental authority, there are good reasons for sovereign states. Justifications for parental authority and allowing parents to be sovereign over certain courses of action for their child include care for children who are unable to care for themselves, access to basic needs (such as food, shelter, health, education, and safety), and the acknowledgment that parents are in a well-suited position to know the needs of their child. Importantly, these justifications for parental authority can surely be overridden and thus the parental authority undermined. Parental authority can be overridden when, for example, the child is placed in imminent danger (such as violence or physical safety) and/or the child is not receiving adequate basic care in the form of, for example, food, shelter, health, and education. Similarly, justifications for state sovereignty include its instrumental value of protecting self-determination as discussed above; but this can be overridden when a state does *not* protect this right on a widespread scale; and, thus, there are violations of human rights on a widespread scale. At this point, the state is no longer serving its citizens by protecting their right to self-determination; and, thus, state sovereignty is no longer justified.

The third feature is that state sovereignty requires an act of recognition. State sovereignty must be recognized to exist. The recognition must be at least reflexive but may also be external. For example, while Taiwan has reflexive recognition and is thus sovereign, *according to itself* – it is missing key external recognition from the international community to be sovereign in the predictable and defensible way that comes from being sovereign, *according to powerful members of the international community*. As the recognition of Taiwan demonstrates, as well as the recognition of state sovereignty present in the ongoing Israeli-Palestinian conflict, not all recognitions of sovereignty carry the same weight (that is, predictable effect and defensibility). For example, a mere reflexive recognition is less valuable than recognition by members of the

UN. The recognition that the international community most cares about is the recognition of the UN.

## THE FORMAL STRUCTURE OF STATE SOVREIGNTY

The following details a formal understanding of how state sovereignty relates to acts of recognition:

(SOV) If state $S$ bears some Sovereign Role, then some state $S_1$ recognizes the sovereignty of $S$

(REC) If state $S_1$ recognizes the sovereignty of state $S$, then $S_1$ publicly documents and declares recognition of the sovereignty of $S$

By logically pinning the Sovereignty Role to *recognition* relations based on documented acts of communication, we can model conflicting claims about state sovereignty and track which claims hold more or less weight. Consider again the example of Taiwan. Instead of immediately entertaining an implied contradiction about the sovereignty of Taiwan, we can exploit SOV and REC thus beginning with *claims* about the sovereignty of Taiwan. For example,

1. Taiwan claims that: "Taiwan is sovereign" [21].
2. China claims that: "Taiwan is not sovereign" [22].

From these claims a user can then assert *recognition relations* between geopolitical entities:

1. Taiwan recognizes the sovereignty of Taiwan
2. China denies the sovereignty of Taiwan

Finally, the user then can decide based on these claims whether the sovereignty of Taiwan should be asserted in their data. In the case where sovereignty is asserted, some associated recognition relation and claim would be logically entailed (from SOV and REC). Hence, the user would relate the sovereignty of Taiwan to the relevant recognition of that sovereignty, like from (3), and the instances of communication, like from (1), where that geopolitical entity expressed recognition of said sovereignty.

Importantly, while this approach allows us to retain all the data without contradiction, it also helps track the weight of various data points (or claims). For example, in this case, while the user may assert in the data that Taiwan is as a matter-of-fact sovereign, the data would reflect that the only associated recognition (in our toy example) of this fact is reflexive, from (3). Were the user to include other data, then Taiwan's sovereignty would be understood in the context of those additional claims, which could be far weightier than the claim in (1).

A benefit of carving out state sovereignty as such, is that state sovereignty can persist in the face of disagreement. One might ask, what happens when a state's sovereignty is recognized by some but explicitly denied by others? The answer is that it depends on who are the affirming and dissenting entities. If, for example, the affirming entities are the UN and entire world order, such as is the case for the USA, Canada, and then UK, then the state's sovereignty will have a great impact on the state's ability to have sufficient authority and independence. On the other hand, if the affirming entity comes only from a reflexive recognition, and the UN does not affirm the state's sovereignty, then the state's sovereignty will be of little consequence. Such is the case for Taiwan. As stated above, to be a sovereign state requires recognition as such; however, not all recognitions are created equal. Importantly, just because some state denies the sovereignty of another state does not mean it will act according to that denial; the converse is also true.

## CONCLUSION

We have outlined first steps towards a formal characterization of state sovereignty, one that is sensitive to UN declarations while also accommodating disagreement and competing claims. The complexities of state sovereignty, as highlighted through examples like Taiwan and Ukraine, demonstrate that the interplay between legal frameworks, international recognition, and political realities, and a full explication of the phenomenon will require careful exploration of each. Much work remains to refine this theory and address the nuances of state sovereignty in an increasingly interconnected and contested global landscape.

REFERENCES

[1] Oppenheim, Lassa F. L. *International Law: A Treatise*. London: Longmans, Green, and Co., 1905.

[2] United Nations Human Rights Council. *Human Rights Situation in the Occupied Palestinian Territory, Including East Jerusalem.* A/HRC/50/21, 2022.

[3] Council on Foreign Relations. "Ukraine: Conflict at the Crossroads of Europe and Russia." *Council on Foreign Relations.* Accessed December 2024.

[4] Republic of China (Taiwan). *Constitution of the Republic of China (Taiwan)*.

[5] National Geospatial-Intelligence Agency. "Taiwan." *Geopolitical Entities, Names, and Codes (GENC) Standard*, version 4.0. U.S. Department of Defense, 2024.

[6] Associated Press. "Paraguay Kicks Out a Visiting Chinese Envoy for Urging Its Lawmakers to Turn Their Backs on Taiwan." *AP News*, December 5, 2024.

[7] United Nations. "Article 4," *Charter of the United Nations*.

[8] United Nations. *Charter of the United Nations*.

[9] United Nations. *Declaration on the Inadmissibility of Intervention in the Domestic Affairs of States and the Protection of Their Independence and Sovereignty*. General Assembly Resolution 2131 (XX), December 21, 1965.

[10] United Nations. *International Covenant on Civil and Political Rights*.

[11] United Nations. "Article 1," *Charter of the United Nations*.

[12] Margalit, Avishai and Joseph Raz, "National Self-Determination," The Journal of Philosophy 87, no. 9 (September 1990): 439–61; Kymlicka, Will. "Minority Rights in Political Philosophy and International Law," in The Philosophy of International Law, ed. Samantha Besson and John Tasioulas (New York, NY: Oxford University Press, 2010), 377–96; Waldron, Jeremy. "Two Conceptions of Self-Determination," in The Philosophy of International Law, ed. Samantha Besson and John Tasioulas (New York, NY: Oxford University Press, 2010), 397–413; Stilz, Anna. Territorial Sovereignty: A Philosophical Exploration (New York, NY: Oxford University Press, 2019).

[13] United Nations. *Vienna Convention on the Law of Treaties*, May 23, 1969.

[14] Philpott, Daniel. "Sovereignty: An Introduction and Brief History." *Journal of International Affairs* 48, no. 2 (Winter 1995): 353–368.

[15] Ibid.

[16] Ibid.

[17] Ibid.

[18] Ibid.

[19] U.S. News & World Report. "A Timeline of the Russia-Ukraine Conflict."

[20] United Nations. "Member States."

[21] Republic of China (Taiwan). *Constitution of the Republic of China.*

[22] Reuters. "China says it takes 'necessary measures' to defend sovereignty over Taiwan." December 11, 2024.